\documentclass[11pt]{article}
\usepackage{bbm}
\usepackage{amsmath,amsfonts,amsthm,mathrsfs}
\usepackage{color}
\usepackage{bm}
\usepackage{amssymb}
\usepackage[dvips]{graphicx}
\usepackage{epsfig}
\usepackage{epsf}
\usepackage{float}
\usepackage{subfigure}
\usepackage{enumerate}
\setlength{\parskip}{0.10in} \textwidth 5.6in \textheight 8.6in
\oddsidemargin .5in \evensidemargin  1in

\newtheorem{example}{Example}
\newtheorem{theorem}{Theorem}
\newtheorem{lemma}{Lemma}
\newtheorem{proposition}{Proposition}
\newtheorem{remark}{Remark}
\newtheorem{corollary}{Corollary}
\newtheorem{definition}{Definition}

\allowdisplaybreaks[4]

\def\dmin{\displaystyle\min}
\def\dmax{\displaystyle\max}
\def\dsum{\displaystyle\sum }
\def\dsup{\displaystyle\sup }
\def\dprod{\displaystyle\prod }
\def\dint{\displaystyle\int }

\def\begeqn{\begin{equation}}
\def\endeqn{\end{equation}}
\def\begth{\begin{theorem}}
\def\endth{\end{theorem}}
\def\begprop{\begin{proposition}}
\def\endprop{\end{proposition}}
\def\begcor{\begin{corollary}}
\def\endcor{\end{corollary}}
\def\begdef{\begin{definition}}
\def\enddef{\end{definition}}
\def\beglemm{\begin{lemma}}
\def\endlemm{\end{lemma}}
\def\begexm{\begin{example}}
\def\endexm{\end{example}}
\def\begrem{\begin{remark}}
\def\endrem{\end{remark}}
\def\beg{\begin}
\def\ga{\alpha}

\def\gb{\beta}

\def\gga{{\gamma}}

\def\gd{\delta}

\def\gep{\varepsilon}

\def\gl{\lambda}

\def\gs{\sigma}

\def\gO{\Omega}

\def\bz{{\bf z}}

\def\N{\mathbb{N}}
\def\R{\mathbb{R}}

\def\X{\mathcal{X}}
\def\Y{\mathcal{Y}}
\def\Z{\mathcal{Z}}
\def\E{\mathcal{E}}

\def\A{\mathcal{A}}

\def\EX{{\mathbb{E}}}

\def\beg{\begin}

\def\gl{\lambda}
\def\R{\mathbb{R}}
\def\cR{\mathcal{R}}

\def\mbS{\mathbb{S}}
\def\tr{\mathbf{\hbox{tr}}}
\def\trace{\mathbf{\hbox{trace}}}

 \beg{document}

 \title{Guaranteed Classification via Regularized Similarity Learning}
\author{Zheng-Chu Guo and Yiming Ying \\ \\
College of Engineering, Mathematics and Physical Sciences\\
University of Exeter, EX4 4QF, UK \\
\{gzhengchu,mathying\}@gmail.com}

\date{}

\maketitle

\begin{abstract}
Learning an appropriate (dis)similarity function from the available data is a central problem in machine learning,  since the success of many machine learning algorithms critically depends on the choice of a similarity function to compare examples. Despite many approaches for similarity metric learning have been proposed, there is little theoretical study on the links between similarity metric learning  and the classification performance of the result classifier. In this paper, we propose a regularized similarity learning formulation associated with general matrix-norms, and  establish their generalization bounds. We show that the generalization error of the resulting linear separator can be bounded by the derived generalization bound of similarity learning. This shows that a good generalization of the learnt similarity function guarantees a good classification of the resulting linear classifier. Our  results extend and improve those obtained by Bellet at~al. \cite{Bellet}. Due to the techniques dependent on the notion of uniform stability \cite{Bous}, the bound obtained there holds true only for the Frobenius matrix-norm regularization. Our techniques using the Rademacher complexity \cite{BM} and its related Khinchin-type inequality enable us to establish bounds for  regularized similarity learning formulations associated with general matrix-norms  including sparse $L^1$-norm and mixed $(2,1)$-norm.
\end{abstract}

\parindent=0cm

\section{Introduction}
The success of many  machine learning algorithms heavily depends on how to specify the {\em similarity} or distance metric between examples.  For instance, the k-nearest neighbor (k-NN) classifier depends on a distance (dissimilarity) function to identify the nearest neighbors for classification.  Most information retrieval methods rely on a similarity function to identify the data points that are most similar to a given query. Kernel methods rely on the kernel function to represent the similarity between examples. Hence, how to learn an appropriate (dis)similarity function from the available data is a central problem in machine learning, which we refer
to as {\em similarity metric learning} throughout the paper.

Recently, a considerable amount of research efforts are devoted to similarity metric learning and many mthods have been proposed. They can be broadly divided into two main categories. The first category of such methods is one-stage approach for similarity metric learning, which means that the methods learn the similarity (kernel) function and classifier together. Multiple kernel learning \cite{lanckrieta:2004,Varma} is a notable one-stage approach, which aims to learn an optimal kernel combination from a prescribed set of positive semi-definite (PSD) kernels. Another exemplary one-stage approach is indefinite kernel learning, which is motivated by the fact that, in many applications,  potential kernel matrices could be non-positive semi-definite. Such cases include hyperbolic tangent kernels \cite{Smola:2000}, and the protein sequence similarity measures derived from Smith-Waterman and BLAST score \cite{Saigo}. Indefinite kernel learning \cite{Chen,YGC} aims to learn a PSD kernel matrix from a prescribed indefinite kernel matrix, which are mostly restricted to the transductive settings.  Recent methods \cite{WuZhou,WuATA} analyzed  regularization networks such as ridge regression and SVM given a prescribed indefinite kernel, instead of aiming to learn an indefinite kernel function from data. The generalization analysis for such one-stage methods was well studied, see e.g. \cite{Chen,YCCOLT,Cortes-a}.

The second category of similarity metric learning is a two-stage method, which means that the processes of learning the similarity function and training the classifier are separate. One exemplar two-stage approach is referred to as  {\em metric learning}  \cite{Bar,Davis,Hoi,Jin,Weinberger,Xing,Ying},  which often focuses on learning a Mahalanobis distance metric defined, for any $x,x'\in \R^d$,  by $d_M(x,x') = \sqrt{(x-x')^T M (x-x')}$. Here, $M$ is a positive semi-definite (PSD) matrix. Another example of such methods \cite{Chechik,Maurer} is {\em bilinear similarity learning}, which focuses on learning a similarity function defined, for any $x, x'\in \R^d,$ by $s_M(x,x')=x^TMx'$ with $M$ being a PSD matrix.  The above methods are mainly motivated by the natural intuition that the similarity score between examples in the same class should be larger than that of examples from distinct classes. The k-NN classification using the similarity metric learnt from the above methods was empirically shown to achieve better accuracy than that using the standard Euclidean distance.

Although many two-stage approaches for similarity metric learning have been proposed, in contrast to the one-stage methods, there is relatively little theoretical study on the question whether  similarity-based learning   guarantees a good generalization of the resultant classification. For instance, generalization bounds were recently established for metric and similarity learning \cite{Cao,Jin,Maurer} under different statistical assumptions on the data.   However, there are no theoretical guarantees for such empirical success. In other words, it is not clear whether good generalization bounds for metric and similarity learning \cite{Jin,Cao} can lead to a good classification performance of the resultant k-NN classifiers. Recently, Bellet et~al. \cite{Bellet} proposed a regularized similarity learning approach, which is mainly motivated by the $(\gep,\gga,\tau)$-good similarity functions introduced in \cite{Balcan2,Balcan}.  In particular, they showed that the proposed similarity learning can theoretically guarantee a good generalization for classification. However, due to the techniques dependent on the notion of uniform stability \cite{Bous}, the generalization bounds only hold true for strongly convex matrix-norm regularization (e.g. the Frobenius norm).

In this paper, we consider a new similarity learning formulation associated with general matrix-norm regularization terms.   Its generalization bounds are established for various matrix regularization including the Frobenius norm, sparse $L^1$-norm, and mixed $(2,1)$-norm (see definitions below). The learnt similarity  matrix is used to design a sparse classification algorithm and we prove the generalization error of its resultant linear separator can be bounded by the derived generalization bound for similarity learning.   This implies that the proposed similarity learning with general matrix-norm regularization guarantees a good generalization for classification.  Our techniques using the Rademacher complexity \cite{BM} and the important Khinchin-type inequality for the Rademacher variables enables us to derive bounds for general matrix-norm regularization including the sparse $L^1$-norm and mixed $(2,1)$-norm regularization.

The remainder of this paper is organized as follows. In Section \ref{sec:main-result}, we propose the similarity learning formulations with general matrix-norm  regularization terms and state the main theorems. In particular, the results will be illustrated using various examples. The related work is discussed in Section \ref{sec:relatedwork}. The generalization bounds for similarity learning are established in Section \ref{sec:generalizationbound}.  In Section \ref{sec:classification}, we develop a theoretical link between the generalization bounds of the proposed similarity learning method and the generalization error of the linear classifier built from the learnt similarity function. Section \ref{sec:rad} estimates the Rademacher averages and gives the proof for examples stated in Section \ref{sec:main-result}.   Section \ref{sec:conclusion} summarizes this paper and discuss some possible directions for future research.

\section{Regularization Formulation and Main Results}\label{sec:main-result}
In this section, we mainly introduce the regularized formulation of similarity learning and state our main results. Before we do that, let us introduce some notations and present some background material.

Denote, for any $n\in\N$, $\N_n=\{1,2,\ldots,n\}.$ Let ${\bf z}=\{z_i=(x_i,y_i): i\in \N_m\}$ be a set of training samples, which is drawn identically  and independently from a distribution $\rho$ on $\Z=\X\times \Y.$ Here, the input space $\X$ is a domain in $\R^d$ and $\Y=\{-1,1\}$ is called the output space. For any $x,x'\in \X,$ we consider $K_A(x,x')=x^TAx'$ as a bilinear similarity score parameterized by a symmetric matrix $A\in \mbS^{d\times d}.$ The symmetry of matrix $A$ guarantees the symmetry of the similarity score $K_A$, i.e.  $K_A(x,x') = K_A(x',x).$

The aim of similarity learning is to learn a matrix $A$ from a given set of training samples ${\bf z}$ such that the similarity score $K_A$ between examples from the same label is larger than that between examples from different labels. A natural approach to achieve the above aim   is to minimize the following empirical error
\begin{equation}\label{eq:hingeloss}
\E_\bz(A) = {1\over m} \dsum_{i\in\N_m}\bigl(1- {1\over mr} \dsum_{j\in\N_m} y_i y_j K_A (x_i,x_j)\bigr)_+,
\end{equation}
where $r>0$ is the margin. Note that $\sum_{j\in\N_m} y_i y_j K_A (x_i,x_j) = \sum_{\{j: y_j = y_i\}} K_A (x_i,x_j)- \sum_{\{j: y_j \neq y_i\}} K_A (x_i,x_j).$ Minimizing the above empirical error encourages, for any $i$, that, with margin $r$, the average similarity scores between examples with the same class as $y_i$ are relatively larger  than those between examples with distinct classes from $y_i$. To avoid overfitting, we add a matrix-regularized term to the above empirical error and reach the following regularization formulation
\begin{equation}\label{algorithm}
A_\bz =
\arg\min_{A\in\mbS^{d\times d}}\Big[ \E_\bz(A) +\lambda\|A\|\Big],
\end{equation}
where $\lambda>0$ is the regularization parameter. Here,  the notation  $\|A\|$ denotes a  general matrix norm. For instance, it can be the sparse $L^1$-norm $\|A\|_1=\sum_{k\in\N_d}\sum_{\ell\in\N_d} |A_{k\ell}|$, the $(2,1)$-mixed norm $\|A\|_{(2,1)}:=\sum_{k\in\N_d}\bigl(\sum_{\ell\in\N_d} A_{k\ell}^2\bigr)^{1\over 2},$ the Frobenius norm $\|A\|_F=\bigl(\sum_{k,\ell\in\N_d}A_{k\ell}^2\bigr)^\frac{1}{2}$ or the trace norm $\|A\|_{\tr}:=\sum_{\ell\in\N_d} \sigma_\ell(A),$ where $\{\sigma_\ell(A): \ell\in \N_d\}$ denote the singular values of matrix $A.$

The first contribution of this paper is to establish generalization bounds for regularized similarity learning (\ref{eq:hingeloss}) with general matrix-norms.  Specifically, define \begeqn\label{eq:trueerror}\E(A) =  \dint_\Z
\left(1-{1\over r}\dint_\Z y y' K_A (x,x')d\rho(x',y')\right)_+ d\rho(x,y).\endeqn The target of generalization analysis for similarity learning is to bound $\E(A_{\bf z})-\E_\bz(A_{\bf z})$.  Its special case with the Frobenius matrix norm was established in \cite{Bellet}. It used the uniform stability techniques  \cite{Bous}, which, however, can not deal with non-strongly convex matrix-norms such as the $L^1$-norm, $(2,1)$-mixed norm and trace norm. Our new analysis techniques are able to deal with general matrix norms, which depend on  the concept of Rademacher averages \cite{BM} defined as follows.
\begin{definition}\label{rademahcer}
Let $\mathcal{F}$ be a class of uniformly bounded functions. For every integer $n$, we call
$$R_n(\mathcal{F}):=\mathbb{E}_{\bf z}\mathbb{E}_{\sigma}\left[\sup_{f\in F}\frac{1}{n}\sum_{i\in\N_n}\sigma_i f(z_i)\right],$$
the  Rademacher average over $\mathcal{F},$ where $\{z_i: i\in\N_n\}$ are independent random variables distributed according to some probability measure and $\{\sigma_i: i\in\N_n\}$ are independent Rademacher random variables, that is, $P(\sigma_i= 1)=P(\sigma_i=-1)=\frac{1}{2}.$
\end{definition}
Before stating our generalization bounds for similarity learning, we first introduce some notations.
For any $B, A\in\R^{n\times d},$ let $\langle B ,A\rangle=\trace(B^TA),$ where $\trace(\cdot)$ denotes the trace of a matrix. For any matrix-norm $\|\cdot\|$, its dual norm $\|\cdot\|_\ast$ is  defined, for any $B$, by $\|B\|_\ast = \sup_{\|A\|\le 1}\trace(B^T A).$ Denote $\|X\|_\ast = \sup_{x,x'\in \X} \|x' x^T\|_\ast$. Let the Rademacher average with respect to the dual matrix norm be defined by
\begeqn\label{eq:Rn}\cR_m : = \EX_{\bz,\sigma}\Big[\dsup_{\tilde{x}\in \X}\Big\|\frac{1}{m}\dsum_{i\in\N_m}\sigma_i y_i x_i \tilde{x}^T \Big\|_{\ast}\Big].\endeqn
Now we can state the generalization bounds for similarity learning, which is closely related to the Rademacher averages with respect to the dual matrix-norm $\|\cdot\|_\ast.$

\begin{theorem}\label{thm:generalresult}
Let $A_\bz$ be the solution to
algorithm (\ref{algorithm}). Then, for any $0<\delta<1,$ with probability at least $1-\gd,$ there holds
\begeqn\label{eq:gen-similarity}\E_\bz(A_\bz)-  \E(A_\bz) \le {6\cR_m\over r\lambda}+ \frac{2X_{*}}{r\lambda}\sqrt{{2\ln \bigl({1\over \gd}\bigr)\over  m}}.\endeqn
\end{theorem}

The proof for Theorem \ref{thm:generalresult} will be given in Section \ref{sec:generalizationbound}. Following the exact argument, similar result is also true if we switch the position of $\E_\bz(A_\bz)$ and $\E(A_\bz)$, i.e. for any $0<\delta<1,$ with probability at least $1-\gd,$ we have
$$ \E(A_\bz) \le \E_\bz(A_\bz) + {6\cR_m\over r\lambda}+ \frac{2X_{*}}{r\lambda}\sqrt{{2\ln \bigl({1\over \gd}\bigr)\over  m}}.
$$

The second contribution of this paper is to investigate the theoretical relationship between similarity learning (\ref{algorithm}) and the generalization error of the linear classifier built from the learnt metric $A_\bz.$ We show that the generalization bound for the similarity learning gives an upper bound for the generalization error of a linear classifier produced by the linear Support Vector Machine (SVM) \cite{Vapnik} defined as follows:
\begin{equation}\label{eq:lsvm}
f_{\bz}=\arg\min\Big\{\frac{1}{m} \sum_{i\in\N_m}\bigl(1-y_if(x_i)\bigr)_+ : ~~f\in\mathcal{F}_{\bf z},  ~\Omega(f):= \sum_{j\in\N_m }|\alpha_j|\leq 1/r\Big\}.
\end{equation}
where $\mathcal{F}_{\bf z}=\Big\{f: f=\sum_{j\in\N_m} \alpha_j K_{A_{\bf z}}(x_j,\cdot),~ a_j\in\mathbb{R}\Big\}$ is the sample-dependent hypothesis space. The empirical error of $f\in\mathcal{F}_\bz$ associated with $\bz$ is defined by
$$\mathscr{E}_\bz(f)=\frac{1}{m}\sum_{i\in\N_m} \bigl(1-y_i f(x_i)\bigr)_+.$$
The true generalization error is defined as
$$\mathscr{E}(f)=\dint_\Z \Big(1-yf(x)\Big)_+ d\rho(x,y).$$
Now we are in a position to state the  relationship between the generalization error  of similarity learning and the generalization error of the liner classifier.
\begin{theorem}\label{thm:mainresult}
Let $A_{\bf z}$ and $f_{\bf z}$ be defined by (\ref{algorithm}) and (\ref{eq:lsvm}), respectively. Then, for any $0<\delta<1,$ with confidence at least $1-\delta,$ there holds
\begeqn\label{eq:relation-inequality}
\mathscr{E}(f_\bz)\leq \E_{\bf z}(A_{\bf z})+
\frac{4\cR_m}{\lambda r} +\frac{2X_*}{\lambda r} \sqrt{\frac{2\log\frac{1}{\delta}}{m}}.\endeqn
\end{theorem}
The proof for Theorem \ref{thm:mainresult} will be established in Section {\ref{sec:classification}}.

Theorems \ref{thm:generalresult} and \ref{thm:mainresult} depend critically on two terms: the constant $X_\ast$ and the Rademacher average $\cR_m.$
Below, we list the estimation of these two terms associated with different matrix norms. For any vector $x = (x^1,x^2,\ldots,x^d)\in \R^d$, denote $\|x\|_\infty = \max_{\ell\in\N_d}|x^\ell|.$

\begin{example}\label{exm:L1} Consider the matrix norm be the sparse $L^1$-norm defined, for any $A\in \mbS^{d\times d}$, by $\|A\| = \sum_{k,\ell\in\N_d} |A_{k\ell}|.$ Let $A_{\bf z}$ and $f_{\bf z}$ be  defined respectively by (\ref{algorithm}) and (\ref{eq:lsvm}). Then, we have the following results.
\begin{enumerate}[(a)]
 \item $X_\ast \le \sup_{x\in \X} \|x\|^2_\infty$ and $\cR_m\le 2\sup_{x\in \X} \|x\|^2_\infty\sqrt{e\log(d+1)\over m}.$
 \item For any $0<\delta<1,$ with confidence at least $1-\delta,$ there holds
 \begeqn\label{eq:exm-1-2}\E_\bz(A_\bz)-  \E(A_\bz) \le {12\sup_{x\in\X}\|x\|^2_\infty\over r\gl}{\sqrt{e\log(d+1)\over m}} + \frac{2\sup_{x\in\X}\|x\|_\infty^2}{r\lambda}\sqrt{{2\ln \bigl({1\over \gd}\bigr)\over  m}}.\endeqn
 \item For any $0<\delta<1,$ with confidence at least $1-\delta,$ there holds
\begeqn\label{eq:exm-1}\mathscr{E}(f_\bz) \leq  \E_{\bf z}(A_{\bf z})+
\frac{4\sup_{x\in\X}\|x\|_\infty^2}{\lambda r}{ \sqrt{2e\log(d+1)\over m}} +\frac{2\sup_{x\in\X} \|x\|_\infty^2}{\lambda r} \sqrt{\frac{2\log\frac{1}{\delta}}{m}}.\endeqn
\end{enumerate}
\end{example}
For any vector $x\in \R^d$, let $\|x\|_F$ be the standard Euclidean norm. Considering the regularized similarity learning with the Frobenius matrix norm, we have the following result.
\begin{example}\label{exm:fro} Consider the Frobenius matrix norm defined, for any $A\in \mbS^{d\times d}$, by $\|A\| = \sqrt{\sum_{k,\ell\in\N_d} |A_{k\ell}|^2}.$ Let $A_{\bf z}$ and $f_{\bf z}$ be defined by (\ref{algorithm}) and (\ref{eq:lsvm}), respectively. Then, we have the following estimation.
\begin{enumerate}[(a)]
\item $X_\ast \le \sup_{x\in \X} \|x\|^2_F$ and $\cR_m\le 2\sup_{x\in \X} \|x\|^2_F \sqrt{1\over m}.$
\item For any $0<\delta<1,$ with confidence at least $1-\delta,$ there holds
\begeqn\label{eq:exm-2-2}\E_\bz(A_\bz)-  \E(A_\bz) \le {6\sup_{x\in\X}\|x\|^2_F\over r\gl \sqrt{m}} + \frac{2\sup_{x\in\X}\|x\|_F^2}{r\lambda}\sqrt{{2\ln \bigl({1\over \gd}\bigr)\over  m}}.\endeqn
\item For any $0<\delta<1,$ with confidence at least $1-\delta,$ there holds
\begeqn\label{eq:exm-2}
\mathscr{E}(f_\bz) \leq  \E_{\bf z}(A_{\bf z})+
\frac{4\sup_{x\in\X}\|x\|_F^2}{\lambda r\sqrt{m}} +\frac{2\sup_{x\in\X}\|x\|_F^2 }{\lambda r} \sqrt{\frac{2\log\frac{1}{\delta}}{m}}.\endeqn
\end{enumerate}
\end{example}
We end this section with two remarks. Firstly, the above theorem and examples mean that a good similarity (i.e. a small generalization error $\E_\bz(A_\bz)$ for similarity learning) can guarantee a good classification (i.e. a small classification error $\mathscr{E}(f_\bz)$). Secondly, the bounds in Example \ref{exm:fro} is consistent with that in \cite{Bellet}. 

\section{Related Work}\label{sec:relatedwork}

In this section, we  discuss studies on similarity metric learning which are related to our work.

Many similarity metric learning methods have been motivated by the intuition that the similarity score between examples in the same class should be larger than that of examples from distinct classes, see e.g. \cite{Bar,Cao,Chechik,Hoi,Jin,Maurer,Weinberger,Xing}.  Jin et~al. \cite{Jin} established generalization bounds for regularized metric learning algorithms via the concept of uniform stability \cite{BM}, which, however, only works for strongly convex matrix regularization terms. A very recent work \cite{Cao} established generalization bounds for the metric and similarity learning associated with general matrix norm regularization using techniques of Rademacher averages and U-statistics. However, there was no theoretical links between the similarity metric learning and the generalization performance of classifiers based on  the learnt similarity matrix. Here, we focused on the problem how to learn a good linear similarity function $K_A$ such that it can guarantee a good classification error of the resultant classifier derived from the learnt similarity function. In addition, our formulation (\ref{algorithm}) is quite distinct from similarity metric learning methods \cite{Cao,Chechik}, since they are based on pairwise or triplet-wise constraints and considered the following pairwise empirical objective function:
\begin{equation}\label{hingeloss1}
\frac{1}{m(m-1)}\sum_{i,j=1,i\ne j}^m\bigl(1-y_iy_j (K_A(x_i,x_j)-r)\bigr)_+.
\end{equation}Our formulation (\ref{algorithm}) is less restrictive since the empirical objective function is defined over an average of similarity scores and it doesn't require the positive semi-definiteness of the similarity function $K.$

Balcan et al. \cite{Balcan} developed a theory of $(\epsilon,\gamma,\tau)$-good similarity function defined as follows. It attempts to investigate the theoretical relationship between the properties of a similarity function and its performance in linear classification.
\begin{definition} \emph{(\cite{Balcan2})}\label{def:goodsimilarity}
A similarity function $K$ is a $(\epsilon,\gamma,\tau)$-good similarity function in hinge loss for a learning problem $P$ if there exists a random indicator function $R(x)$ defining a probabilistic set of
``reasonable points'' such that the following conditions hold:\\
1. $\mathbb{E}_{(x,y)\sim P}[1-yg(x)/\gamma]_+\leq \epsilon,$ {\rm where} $g(x)=\mathbb{E}_{(x',y')\sim P}[y'K(x,x')|R(x')],$\\
2. ${\rm Pr}_{x'}[R(x')]\ge \tau.$
\end{definition}
The first condition can be interpreted as ``most points $x$ are on average $2\gamma$ more similar to random reasonable points of the same
class than to random reasonable points of the distinct classes'' and the second condition as ``at least a
$\tau$ proportion of the points should be reasonable.''  The following theorem implies that if given an $(\epsilon,\gamma,\tau)$-good similarity function and enough landmarks, there exists a separator $\alpha$ with error arbitrarily close to $\epsilon.$
\begin{theorem} \emph{(\cite{Balcan2})}\label{thm:Balcan} Let $K$ be an $(\epsilon,\gamma,\tau)$-good similarity function in hinge loss for a learning problem $P.$ For any $\epsilon_1>0,$ and $0<\delta\leq \gamma\epsilon_1/4,$ let $S=\{x_1', \cdots, x_{d_{land}}'\}$ be a potentially unlabeled sample of $d_{land}=\frac{2}{\tau}\big(\log(2/\delta)+16\frac{\log(2/\delta)}{(\epsilon_1\gamma)^2}\big)$ landmarks drawn from $P.$ Consider the mapping $\phi_i^S=K(x,x_i'),$ $i\in \{1,\cdots,d_{land}\}.$ Then, with probability at least $1-\delta$ over the random sample $S,$ the induced distribution $\phi^S(P)$ in $\R^{d_{land}}$ has a linear separator $\alpha$ of error at most $\epsilon+\epsilon_1$ at margin $\gamma.$
\end{theorem}
It was mentioned in \cite{Balcan} that the linear separator  can be estimated by solving the following linear programming if we have $d_u$ potentially unlabeled sample and $d_l$ labeled sample,
\begin{equation}\label{equ:optimization}
\min_\ga\Big\{\sum_{i=1}^{d_l}\Big[1-\sum_{j=1}^{d_u} \alpha_j y_i K(x_i,x_j')\Big]_+: ~~ \sum_{j=1}^{d_u}|\alpha_j|\leq 1/\gamma\Big\}.
\end{equation}
The above   algorithm (\ref{equ:optimization}) is quite similar to the linear SVM (\ref{eq:lsvm}) used in our paper. Our work is distinct from Balcan et al. \cite{Balcan} in the following two aspects. Firstly, the similarity function $K$ is predefined in algorithm (\ref{equ:optimization}), while we aim to learn a similarity function $K_{A_\bz}$ from a regularized similarity learning formulation (\ref{algorithm}). Secondly, although the separators are both trained from the linear SVM,  the classification algorithm (\ref{equ:optimization}) in \cite{Balcan}  was designed using two different sets of examples, a set of labeled samples of size $d_l$ to train the classification algorithm and another set of unlabeled samples with size $d_u$ to define the mapping $\phi^S.$ In this paper, we used the same set of training samples for both similarity learning (\ref{algorithm}) and the classification algorithm (\ref{eq:lsvm}).

Recent work by Bellet et~al. \cite{Bellet} is mostly close to ours. Specifically, they considered similarity learning formulation (\ref{algorithm}) with the Frobenius norm regularization. Generalization bounds for similarity learning were derived via uniform stability arguments \cite{Bous} which can not deal with, for instance, the $L^1$-norm and $(2,1)$-norm regularization terms.  In addition, the results about the relationship between the similarity learning and the performance of the learnt matrix in classification were quoted from \cite{Balcan} and hence requires two separate sets of samples to train the classifier.

Most recently, there is a considerable interest on two-stage  approaches for multiple kernel learning \cite{Cortes-b,Kar} which perform competitively as  the one-stage approaches \cite{lanckrieta:2004,Varma}. In particular,  Kar \cite{Kar} studied generalization guarantees for the following regularization formulation for learning similarity (kernel) function:  \begin{equation}\label{mkl-mu}\arg\dmin_{\mu\ge 0}
{2\over m(m-1)}\dsum_{1\le i<j\le n }^m\bigl(1-y_iy_j K_\mu(x_i,x_j)\bigr)_+  +  \Omega(\mu).
\end{equation}
where $K_\mu = \sum_{\ell=1}^p \mu_\ell K_\ell$ is the positive linear combination of base kernels $\{K_\ell: \ell=1,2,\ldots,p\},$ and $\Omega(\cdot)$ is a regularization term which, for instance, can be the Frobenius norm or the $L^1$ norm. Specifically, Kar  \cite{Kar} established elegant generalization bounds for the above two-stage multiple kernel learning using techniques of Rademacher averages  \cite{BM,Kakade-a,Kakade-b} and U-statistics \cite{Cao,Clem}.  The empirical error term (\ref{eq:hingeloss}) in our formulation (\ref{algorithm}) is not a U-statistics term and the techniques in \cite{Kar,Cao} can not directly be applied to our case.

Jain et~al. \cite{Jain} and Kar et~al. \cite{Kar} introduced an extended framework of \cite{Balcan2,Balcan} in the general setting of supervised learning. The authors proposed a general goodness criterion for similarity functions, which can handle general supervised learning tasks and also subsumes the goodness of condition of \cite{Balcan}. There, efficient algorithms were constructed with provable generalization error bounds. The main distinction between these work and our work is that we aim to learn a similarity function while in their work a similarity function is defined in advance.

\section{Generalization Bounds for Similarity Learning}\label{sec:generalizationbound}
In this section, we establish generalization bounds for the similarity learning formulation (\ref{algorithm}) with general matrix-norm regularization terms. Recall that the true error for similarity learning is defined by
$$\E(A) =  \dint_\Z
\Big(1-{1\over r}\dint_\Z y y' K_A (x,x')d\rho(x',y')\Big)_+d\rho(x,y).$$
The target of generalization analysis for similarity learning is to bound the true error $\E(A_{\bf z})$ by the empirical error $\E_\bz(A_{\bf z}).$

By the definition (\ref{algorithm}) of $A_\bz$, we know that $ \E_\bz(A_\bz) + \gl \|A_\bz\| \le  \E_\bz(0) + \gl \|0\| = 1,$
which implies that $ \|A_\bz\|\le {1/\gl}.$ Denote
$$ \A = \Bigl\{A\in\mbS^{d\times d}:  \|A\| \le {1/\gl} \Bigr\}. $$
Hence, one can easily see that the solution $A_\bz$ to algorithm (\ref{algorithm}) belongs to  $\mathcal{A}.$ Now we are ready to prove generalization bounds for similarity learning which was stated as Theorem \ref{thm:generalresult} in Section \ref{sec:main-result}.

\noindent {\bf Proof of Theorem \ref{thm:generalresult}:} Our proof is divided into two steps.

{\bf Step 1}: ~ Let $\EX_\bz$ denote the expectation with respect
to samples $\bz$. Observe that $\E_{{\bz}}(A_\bz) -
\E(A_\bz) \le \dsup_{A\in \A} \Bigl[\E_\bz(A) -
\E(A)\Bigr].$ Also, for any ${\bf z} = (z_1,\ldots, z_k,\ldots,
z_m)$ and $ \tilde{\bz}=(z_1,\ldots, \tilde{z}_k,\ldots, z_m),$ $1\leq k\leq m$, there holds
$$\begin{array}{ll}& \Bigl|\dsup_{A\in \A} \Bigl[\E_\bz(A) -
\E(A)\Bigr] -\dsup_{A\in \A} \Bigl[\E_{\tilde{\bz}}(A) -
\E(A)\Bigr] \Bigr| \le \dsup_{A\in \A}|\E_\bz(A)
-  \E_{\tilde{\bz}}(A)| \\ &\leq{1\over m^2 r}\dsup_{A\in
\A}\Big\{\dsum_{i=1,i\neq
k}^m|y_i y_k K_A(x_k,x_i) - y_i \tilde{y}_k K_A(\tilde{x}_k,x_i)|\\
&~~~+ |\sum_{j\in\N_m} (y_k y_j K_A(x_k,x_j)-\tilde{y}_k y_j K_A(\tilde{x}_k,x_j))|  \Big\}\\
&\le  {2\over m^2 r}\dsup_{A\in
\A}\dsum_{i\in\N_m}
\Big(|y_i y_k K_A(x_k,x_i)| + |y_i \tilde{y}_k K_A(\tilde{x}_k,x_i)|\Big)\leq \frac{4 X_{*}}{mr\lambda}.
\end{array}$$
Applying the McDiarmid's inequality \cite{McD} (see Lemma \ref{lem:McD}
in the Appendix) to the term $\dsup_{A\in \A} \Bigl[\E_\bz(A) -
\E(A)\Bigr]$, with probability at least $1-{\gd},$ there holds
\begeqn\begin{array}{ll}\label{eq:Mcd-1} \dsup_{A\in \A}
\Bigl[\E_\bz(A) - \E(A)\Bigr]& \le \EX_\bz\dsup_{A\in \A}
\Bigl[\E_\bz(A) - \E(A)\Bigr]  + \frac{2X_{*}}{r\lambda}\sqrt{{2\ln \bigl({1\over \gd}\bigr)\over  m}}.
\end{array}\endeqn

Now we are in a position to estimate the first term in the
expectation form on the righthand side of the above equation by
standard symmetrization techniques.

{\bf Step 2}: ~ We divide the term $\EX_\bz\dsup_{A\in \A}
\Bigl[\E_\bz(A) - \E(A)\Bigr]$ into two parts as follows, $$\begin{array}{ll}
&\EX_\bz\dsup_{A\in \A}
\Bigl[\E_\bz(A) - \E(A)\Bigr]\\
&= \EX_\bz\dsup_{A\in \A}\Big\{ {1\over m} \dsum_{i\in\N_m}\Bigl[1- {1\over mr} \dsum_{j\in\N_m} y_i y_j K_A (x_i,x_j)\Bigr]_+ - \E(A)  \Big\}\\
&= \EX_\bz\dsup_{A\in \A}\Big\{\frac{1}{m}\dsum_{i\in\N_m}\Bigl[1-{1\over r}\EX_{(x',y')}y_i y'K_A(x_i,x')\Bigr]_+ -\E(A)\\
&~~~- {1\over m} \dsum_{i\in\N_m}\Bigl[1-{1\over r}\EX_{(x',y')}y_i y'K_A(x_i,x')\Bigr]_+ +{1\over m} \dsum_{i\in\N_m}\Bigl[1- {1\over mr} \dsum_{j\in\N_m} y_i y_j K_A (x_i,x_j)\Bigr]_+ \Big\}\\
& \leq I_1 + I_2,
\end{array}$$
where
$$I_1:=\EX_\bz\dsup_{A\in \A}\Big\{\frac{1}{m}\dsum_{i\in\N_m}\Bigl[1-{1\over r}\EX_{(x',y')}y_i y'K_A(x_i,x')\Bigr]_+ - \E(A) \Big\},$$ and
$$I_2:= -\EX_\bz\dsup_{A\in \A}\Big\{\frac{1}{m}\dsum_{i\in\N_m}\Bigl[1-{1\over r}\EX_{(x',y')}y_i y'K_A(x_i,x')\Bigr]_++{1\over m} \dsum_{i\in\N_m}\Big(1- {1\over mr} \dsum_{j\in\N_m} y_i y_j K_A (x_i,x_j)\Big)_+ \Big\}.$$
Now let
$\bar{\bz} = \{\bar{z}_1,\bar{z}_2,\ldots, \bar{z}_m\}$  be an i.i.d. sample which is independent of $\bz.$ We first estimate $I_1$ using the standard symmetrization techniques, to this end, we rewrite $\E(A)$ as $\EX_{\bar{\bz}}\Big(\frac{1}{m}\dsum_{i\in\N_m}\Bigl[1-{1\over r}\EX_{(x',y')}\bar{y}_i y'K_A(\bar{x}_i,x')\Bigr]_+\Big)$. Then we have
$$\begin{array}{ll}
I_1 &=\EX_\bz\dsup_{A\in \A}\Big\{\frac{1}{m}\dsum_{i\in\N_m}\Bigl[1-{1\over r}\EX_{(x',y')}y_i y'K_A(x_i,x')\Bigr]_+\\ &~~~- \EX_{\bar{\bz}}\Big(\frac{1}{m}\dsum_{i\in\N_m}\Bigl[1-{1\over r}\EX_{(x',y')}\bar{y}_i y'K_A(\bar{x}_i,x')\Bigr]_+\Big\} \\
&\leq \EX_{\bz,{\bf \bar{z}}} \dsup_{A\in \A}\Big\{\frac{1}{m}\dsum_{i\in\N_m}\Bigl[1-{1\over r}\EX_{(x',y')}y_i y'K_A(x_i,x')\Bigr]_+ \\ &~~~- \frac{1}{m}\dsum_{i\in\N_m}\Bigl[1-{1\over r}\EX_{(x',y')}\bar{y}_i y'K_A(\bar{x_i},x')\Bigr]_+ \\
\end{array}$$
By the standard Rademacher symmetrization technique and  the contraction property of the Rademacher average (see Lemma \ref{lem:contr-prop} in the Appendix), we further have
$$\begin{array}{ll}
I_1&\leq 2\EX_{\bz,\sigma} \dsup_{A\in \A}\Big\{ \frac{1}{m}\dsum_{i\in\N_m} \sigma_i\Bigl[1-{1\over r}\EX_{(x',y')}y_i y'K_A(x_i,x')\Bigr]_+\Big\}\\
&\leq 4\EX_{\bz,\sigma} \dsup_{A\in \A}\Big| \langle\frac{1}{mr}\dsum_{i\in\N_m}\sigma_i y_i x_i\int y'x'^T d\rho(x',y'),A \rangle\Big|\\
&\leq {4\over r\lambda}\EX_{\bz,\sigma} \Big\|\frac{1}{m}\dsum_{i\in\N_m}\sigma_i y_i x_i\int y'x'^T d\rho(x',y')\Big\|_{*}\leq {4\over r\lambda}\EX_{\bz,\sigma} \dsup_{\tilde{x}}\Big\|\frac{1}{m}\dsum_{i\in\N_m}\sigma_i y_i x_i \tilde{x}^T \Big\|_{*},
\end{array}$$
where the last inequality follows from the fact that $\langle A,B\rangle\leq \|A\|\|B\|_{*}\leq \frac{1}{r}\|B\|_{*}$ for any $A\in\mathcal{A}$ and $B\in\R^{d\times d}.$

Similarly, we can estimate $I_2$ as follows.
$$\begin{array}{ll}
I_2&=\EX_\bz\dsup_{A\in \A}\frac{1}{m}\dsum_{i\in\N_m}\Big(\Bigl[1- {1\over mr} \dsum_{j\in\N_m} y_i y_j K_A (x_i,x_j)\Bigr]_+  - \Bigl[1-{1\over r}\EX_{(x',y')}y_i y'K_A(x_i,x')\Bigr]_+  \Big)\\
&\leq \EX_\bz\dsup_{A\in \A}\frac{1}{m}\dsum_{i\in\N_m}\Big(\Big|{1\over r}\EX_{(x',y')}y_i y'K_A(x_i,x')-{1\over mr} \dsum_{j\in\N_m} y_i y_j K_A (x_i,x_j)\Big| \Big)\\
&= \EX_\bz\dsup_{A\in \A}\frac{1}{mr}\dsum_{i\in\N_m}\Big(\Big|\langle \EX_{(x',y')} y_i y'x'x_i^T -\frac{1}{m}\dsum_{j\in\N_m} y_i y_j x_j x_i^T,A \rangle \Big|\Big)\\
&\leq \frac{1}{r\lambda}\EX_\bz \dsup_{x\in\X}\Big\| \EX_{(x',y')} y'x'x^T -\frac{1}{m}\dsum_{j\in\N_m}  y_j x_j x^T \Big\|_{*}\\
&= \frac{1}{r\lambda}\EX_\bz\dsup_{x\in\X}\Big\|\EX_{\bz '}{1\over m}\dsum_{j\in\N_m}y_j'x_j'x^T-\frac{1}{m}\dsum_{j\in\N_m}  y_j x_j x^T \Big\|_{*}.\end{array}$$
In the above estimation, the first inequality follows from the Lipschitz continuity of the hinge loss function. Following the standard Rademacher symmetrization technique (see e.g. \cite{BM}), from the above estimation we can further estimate $I_2$ as follows:
$$\begin{array}{ll}I_2 &\leq \frac{1}{r\lambda}\EX_\bz\dsup_{x\in\X}\Big\|\EX_{\bz '}{1\over m}\dsum_{j\in\N_m}y_j'x_j'x^T-\frac{1}{m}\dsum_{j\in\N_m}  y_j x_j x^T \Big\|_{*}\\
&\leq \frac{1}{r\lambda}\EX_{\bz,{\bf {z'}}}\dsup_{x\in\X}\Big\|{1\over m}\dsum_{j\in\N_m}y_j'x_j'x^T-\frac{1}{m}\dsum_{j\in\N_m}  y_j x_j x^T \Big\|_{*} \\
&\leq \frac{1}{r\lambda}\EX_{\bz,{\bf {z'}},\sigma}\dsup_{x\in\X}\Big\|{1\over m}\dsum_{j\in\N_m}\sigma_j\Big(y_j'x_j'x^T- y_j x_j x^T \Big)\Big\|_{*} \leq \frac{2}{r\lambda}\EX_{\bz,\sigma}\dsup_{x\in\X}\Big\|{1\over m}\dsum_{j\in\N_m}\sigma_j y_j x_j x^T \Big\|_{*}.
\end{array}$$
The desired result follows by combining (\ref{eq:Mcd-1}) with the above estimation for $I_1$ and $I_2.$ This completes the proof for the theorem.
\hfill $\Box$

\section{Guaranteed Classification Via Good Similarity}\label{sec:classification}

In this section, we investigate the theoretical relationship between the generalization error for the similarity learning and that of the linear classifier built from the learnt similarity metric $K_{A_\bz}.$ In particular, we will show that the generalization error of the similarity learning gives an upper bound for the generalization error of the linear classifier which was stated as Theorem \ref{thm:mainresult} in Section \ref{sec:main-result}.

Before giving the proof of Theorem \ref{thm:mainresult}, we first establish the generalization bounds for the linear SVM algorithm (\ref{eq:lsvm}). Recalling that the linear SVM algorithm (\ref{eq:lsvm}) was defined by $$
f_{\bz}=\arg\min\Big\{\frac{1}{m} \sum_{i\in\N_m}\bigl(1-y_if(x_i)\bigr)_+ : ~~f\in\mathcal{F}_{\bf z},  ~\Omega(f):= \sum_{j\in\N_m} |\alpha_j|\leq 1/r\Big\},
$$
where $$\mathcal{F}_{\bf z}=\Big\{f: f=\sum_{j\in\N_m} \alpha_j K_{A_{\bf z}}(x_j,\cdot), a_j\in\mathbb{R}\Big\}.$$ The generalization analysis of the linear SVM algorithm (\ref{eq:lsvm}) aims to estimate the term   $\mathscr{E}(f_\bz)-\mathscr{E}_\bz(f_\bz).$
For any $\bz$, one can easily see that the solution to algorithm (\ref{eq:lsvm}) belongs to the set $\mathcal{F}_{\bz, r},$ where
$$\mathcal{F}_{\bz, r}=\Big\{f=\sum_{j\in\N_m} \alpha_j K_{A_{\bf z}}(x_j,\cdot):  ~~\Omega(f)= \sum_{j\in\N_m} |\alpha_j|\leq 1/r, a_j\in\mathbb{R}\Big\}.$$
To perform the generalization analysis, we seek a sample-independent set which contains, for any $\bz,$ the sample-dependent hypothesis space $\mathcal{F}_{\bf z}.$ Specifically, we define a sample independent hypothesis space by
$$\mathcal{F}_{m}=\Big\{f=\sum_{i\in\N_m} \alpha_i K_{A}(u_i,\cdot):~~ \|A\|\leq 1/\lambda, u_j\in X, a_j\in\mathbb{R}\Big\}.$$
Recalling that, for any $\bz$,  $\|A_\bz \|\le \gl^{-1}$, one can easily see that $\mathcal{F}_{\bf z}$ is a subset of $\mathcal{F}_{m}.$ It follows that, for any $\bz$, the solution to the linear SVM algorithm (\ref{eq:lsvm}) lies in the set $\mathcal{F}_{m, r},$  which is given by
$$\mathcal{F}_{m,r}=\Big\{f\in\mathcal{F}_m: ~~\Omega(f)\leq 1/r\Big\}.$$

The following theorem states the generalization bounds of the linear SVM for classification.
\begin{theorem}\label{thm:linearsvm}
Let $f_{\bf z}$ be the solution to the algorithm (\ref{eq:lsvm}). For any $0<\delta<1,$ with probability at least $1-\delta,$ we have
\begin{equation}\mathscr{E}(f_\bz)-\mathscr{E}_{\bf z}(f_\bz)
\leq \frac{4\cR_m}{\lambda r} +\frac{2X_*}{\lambda r} \sqrt{\frac{2\log\frac{1}{\delta}}{m}}.\end{equation}
\end{theorem}

\begin{proof}
By McDiarmid's inequality, for any $0<\delta<1,$ with confidence $1- \delta$, there holds
$$\begin{array}{ll}&\mathscr{E}(f_\bz)-\mathscr{E}_{\bz}(f_\bz)\leq\dsup_{f\in\mathcal{F}_{\bz,r}}\big(\mathscr{E}(f)-\mathscr{E}_{\bf z}(f)\big)\leq\dsup_{f\in\mathcal{F}_{m,r}}\big(\mathscr{E}(f)-\mathscr{E}_{\bf z}(f)\big)\\
&\leq \EX_{\bf z}\dsup_{f\in\mathcal{F}_{m,r}}\big(\mathscr{E}(f)-\mathscr{E}_{\bf z}(f)\big)+\frac{2X_*}{\lambda r} \sqrt{\frac{2\log\frac{1}{\delta}}{m}}.\end{array}$$
Next, all we need is to estimate the first part of the right hand-side of the above inequality. Let ${\bf\bar{z}}$ be an independent sample (independent each other and $\bz$) and with the same distribution as $\bz.$
$$\begin{array}{ll}&\EX_{\bf z}\dsup_{f\in\mathcal{F}_{m,r}}\big(\mathscr{E}(f)-\mathscr{E}_{\bf z}(f)\big)
=\EX_{\bf z}\dsup_{f\in\mathcal{F}_{m,r}}\big(\EX_{\bf\bar{z}}\mathscr{E}_{\bf \bar{z}}(f)-\mathscr{E}_{\bz}(f)\big)\leq \EX_{\bf z,\bar{z}}\dsup_{f\in\mathcal{F}_{m,r}}\big(\mathscr{E}_{\bf \bar{z}}(f)-\mathscr{E}_{\bf z}(f)\big)\\
&\leq 2\EX_{\bf z,\sigma}\Big[\dsup_{\|A\|\leq 1/\lambda}~\dsup_{\sum_{i\in\N_m} |\alpha_i|\leq 1/r}\Big(\frac{1}{m}\sum_{i\in\N_m} \sigma_{i} \big  [1-\sum_{j\in\N_m}\alpha_j y_i K_{A}(x_i,u_j)]_+\Big)\Big]\\
&\leq 4\EX_{\bf z,\sigma}\Big[\dsup_{\|A\|\leq 1/\lambda}~\dsup_{\sum_{i\in\N_m} |\alpha_i|\leq 1/r}\Big(\frac{1}{m}\sum_{i\in\N_m} \sigma_{i}   \sum_{j\in\N_m}\alpha_j y_i K_{A}(x_i,u_j)\Big)\Big]\\
&\leq \frac{4}{r}\EX_{\bf z,\sigma}\dsup_{A: \|A\|\leq 1/\lambda}~\dsup_{x\in \X}\Big(\Big|\frac{1}{m}\sum_{i\in\N_m} \sigma_{i}   y_i \langle x_i x^T, A\rangle\Big|\Big)\leq  \frac{4}{\lambda r}\EX_{\bf z,\sigma}\dsup_{x\in \X}\Big\|\frac{1}{m}\sum_{i\in\N_m} \sigma_{i}  y_i x_ix^T\Big\|_*.\end{array}$$
Here we also use the standard Rademacher Symmetrization technique and the contractor property of the Rademacher average. Then the proof is completed.
\end{proof}

Now we are in a position to give the detailed proof of Theorem \ref{thm:mainresult}.

\noindent{\bf Proof of Theorem \ref{thm:mainresult}:} If we take $\alpha^0=(\frac{y_1}{mr},\cdots,\frac{y_m}{mr})^T,$ then $f_{\bf z }^0=\frac{1}{mr}\sum_{j\in\N_m} y_j K_{A_{\bf z}}(x_j,\cdot).$ One can easily see that $\Omega(f_{\bf z }^0)=\sum_{j\in\N_m}|\alpha^0_j|=\frac{1}{r},$ that means $f_{\bf z }^0\in\mathcal{F}_{\bz, r}.$ From Theorem \ref{thm:linearsvm} and the definition of $f_\bz$ , we get
$$\begin{array}{ll}\mathscr{E}(f_\bz)&\leq \mathscr{E}_{\bf z}(f_\bz)+
\frac{4 \cR_m}{\lambda r}   +\frac{2X_*}{\lambda r} \sqrt{\frac{2\log\frac{1}{\delta}}{m}} \leq \mathscr{E}_{\bf z}(f_{\bf z }^0)+
\frac{4\cR_m}{\lambda r}  +\frac{2X_*}{\lambda r} \sqrt{\frac{2\log\frac{1}{\delta}}{m}}\\
&= {1\over m} \dsum_{i\in\N_m}\Big(1- {1\over mr} \dsum_{j\in\N_m} y_i y_j K_{A_{\bf z}} (x_i,x_j)\Big)_+ +
\frac{4\cR_m}{\lambda r}  +\frac{2X_*}{\lambda r} \sqrt{\frac{2\log\frac{1}{\delta}}{m}}\\
&=  \E_{\bf z}(A_{\bf z})+
\frac{4 \cR_m}{\lambda r}  +\frac{2X_*}{\lambda r} \sqrt{\frac{2\log\frac{1}{\delta}}{m}}.\end{array} $$
This completes the proof of the theorem.
\hfill $\Box$

\section{Estimating Rademacher Averages}\label{sec:rad}
The main theorems above critically depend on the estimation of the Rademacher average $\cR_m$ defined by equation (\ref{eq:Rn}). In this section, we establish a self-contained proof for this estimation and prove the examples listed in Section \ref{sec:main-result}.  For notational simplicity, denote by $x_i^\ell$ the $\ell$-th variable of the $i$-th sample $x_i\in \R^d.$

\noindent{\bf Proof of Example \ref{exm:L1}:} The dual norm of $L^1$-norm is the $L^\infty$-norm. Hence, \begeqn\label{eq:Xast-L1}X_\ast = \sup_{x,x'\in \X} \sup_{\ell,k\in\N_d}|x^\ell (x')^k| = \sup_{x\in\X} \|x\|^2_\infty.\endeqn Also, the Rademacher average can be rewritten as
\begeqn\label{eq:inter-cR}\cR_m = \EX_{\bz,\sigma}\sup_{x\in\X} \|{1\over m}\dsum_{j\in\N_m}\sigma_j y_j x_j x^T \|_\infty\le \sup_{x\in\X} \|x\|_\infty \; \EX_{\bz,\sigma}\dmax_{\ell\in\N_d} \Big|\frac{1}{m}\dsum_{j\in\N_m}\sigma_j y_j x^\ell_j\Big|. \endeqn
Now let $U_\ell(\sigma)=\frac{1}{m}\dsum_{j\in\N_m}\sigma_j y_j x_j^\ell,$ for any $\ell\in\N_d.$  By Jensen's inequality, for any $\eta>0$, we have
\begeqn\label{eq:uell}\begin{array}{ll}
&e^{\eta^2(\EX_\sigma\max_{\ell\in\N_d} |U_\ell(\sigma)|)^2}-1\leq \EX_\sigma[e^{\eta^2 (\max_{\ell\in\N_d} |U_\ell(\sigma)|)^2}-1]\\
&= \EX_\sigma[\dmax_{\ell\in\N_d}e^{\eta^2  |U_\ell(\sigma)|^2}-1]
\leq\dsum_{\ell\in\N_d} \EX_\sigma[e^{\eta^2 ( |U_\ell(\sigma)|)^2}-1].
\end{array}\endeqn
Furthermore, for any $\ell\in\N_d,$ there holds
$$\begin{array}{ll}
&\EX_\sigma[e^{\eta^2 ( |U_\ell(\sigma)|)^2}-1]=\dsum_{k\geq1}\frac{1}{k!}\eta^{2k}\EX_\sigma|U_\ell|^{2k}\\
&\leq \dsum_{k\geq1}\frac{1}{k!}\eta^{2k} (2k-1)^k (\EX_\sigma|U_\ell|^{2})^k\leq \dsum_{k\geq1}(2e\eta^{2} \EX_\sigma|U_\ell|^{2})^k,
\end{array}$$ where the first inequality follows from the Khinchin-type inequality (see Lemma \ref{Khinchin} in the Appendix), and
the second inequality holds due to the Stirling's inequality:$e^{-k}k^k\leq k!$.
Now set $\eta=[2\sqrt{e}\dmax_{\ell\in\N_d}(\EX_\sigma|U_\ell|^{2})^{1\over2}]^{-1}$. The above inequality can be upper bounded by $$\EX[e^{\eta^2 ( |U_\ell(\sigma)|)^2}-1]\leq \dsum_{k\geq1}2^{-k}=1, ~\forall \ell\in \N_d.$$
Putting the above estimation back into (\ref{eq:uell}) implies that
$$e^{\eta^2(\EX\max_{\ell\in\N_d} |U_\ell(\sigma)|)^2}-1\leq d.$$
That means
\begeqn\label{eq:Rn-L1}\begin{array}{ll}
&\EX_\sigma\dmax_{\ell\in\N_d} |U_\ell(\sigma)|=\EX_\sigma\dmax_{\ell\in\N_d}\bigl|\frac{1}{m}\dsum_{j\in\N_m}\sigma_j y_j x_j^\ell\bigr|\leq \sqrt{\log(d+1)\eta^{-2}}
\\ & =2\sqrt{e\log(d+1)}\dmax_{\ell\in\N_d}(\EX_\sigma|U_\ell|^{2})^{1\over 2}\\
&=2\sqrt{e\log(d+1)}\dmax_{\ell\in\N_d}\Big(\EX_\sigma\Big|\frac{1}{m}\dsum_{j\in\N_m}^n\sigma_j y_j x_j^\ell\Big|^{2}\Big)^{1\over 2}\\
&=2\sqrt{e\log(d+1)}\dmax_{\ell\in\N_d}\Big(\EX_\sigma\frac{1}{m^2}\dsum_{j,k\in\N_m}\sigma_j \sigma_k y_j y_k x_j^\ell x_k^\ell\Big)^{1\over 2}\\
&=2\sqrt{e\log(d+1)}\dmax_{\ell\in\N_d}\Big(\frac{1}{m^2}\dsum_{j\in\N_m}(x_j^\ell)^2 \Big)^{1\over 2}\leq 2\dsup_{x\in\X}\|x\|_\infty{\sqrt{e\log(d+1)\over m}}.
\end{array}\endeqn
Putting the above estimation back into (\ref{eq:inter-cR}) implies that
$$\cR_m   \le 2\dsup_{x\in\X}\|x\|^2_\infty{\sqrt{e\log(d+1)\over m}}. $$
The other desired results in the example follow directly from combining the above estimation with Theorems \ref{thm:generalresult} and \ref{thm:mainresult}.
\hfill $\Box$

We turn our attention to similarity learning formulation (\ref{algorithm}) with the Frobenius norm regularization.

\noindent {\bf Proof of Example \ref{exm:fro}:} The dual norm of the Frobenius norm is itself. Consequently, $X_{*}=\dsup_{x,x'\in\X}\|x'x^T\|_F=\dsup_{x\in\X}\|x\|_F^2.$  The Rademacher average can be rewritten as  $$\cR_m = \EX_{{\bf z},\sigma}\dsup_{x\in\X} \Big\|\frac{1}{m}\dsum_{j\in\N_m}\sigma_j y_j x_j x^T\Big\|_{F}.$$  By Cauchy's Inequality, there holds
\begeqn\label{eq:Rn-Fro}\begin{array}{ll}\cR_m & =\EX_{{\bf z},\sigma}\dsup_{x\in\X} \|x\|_F \Big\|\frac{1}{m}\dsum_{j\in\N_m}\sigma_j y_j x_j \Big\|_F \\ & \leq \dsup_{x\in\X} \|x\|_F  \EX_{\bf z}\bigl(\EX_{\sigma} \big\|\frac{1}{m}\dsum_{j\in\N_m}\sigma_j y_j x_j \big\|^2_F\bigr)^{1\over 2}   \\ &
= \dsup_{x\in\X} \|x\|_F \EX\bigl(\sum_{j\in\N_m} \|x_j\|_F^2\bigr)^{1\over 2 }\big / m\leq \dsup_{x\in\X} \|x\|_F^2 {1\over \sqrt{m}}.\end{array}\endeqn
Then, the desired results can be derived by combining the above estimation with Theorems \ref{thm:generalresult} and \ref{thm:mainresult}.
\hfill $\Box$

The above generalization bound for similarity learning formulation (\ref{algorithm}) with the Frobenius norm regularization is consistent with that given in \cite{Bellet}, where the result holds true under the assumption that $\sup_{x\in \X} \|x\|_F\leq 1$. Below, we provide the estimation of $\cR_m$ respectively for the mixed $(2,1)$-norm and the trace norm.
\begexm\label{exm:21norm} Consider similarity learning formulation (\ref{algorithm}) with the mixed $(2,1)$-norm regularization $\|A\|_{(2,1)} = \sum_{k\in\N_d}(\sum_{\ell\in\N_d} |A_{k\ell}|^2)^{1/2}.$ Then, we have the following estimation.

\beg{enumerate}[(a)]
\item $X_\ast \le \bigl[\sup_{x\in \X}\|x\|_F\bigr]\bigl[\sup_{x\in \X}\|x\|_\infty\bigr]$ and $$\cR_m\le 2\bigl[\sup_{x\in \X}\|x\|_F\bigr]\bigl[\sup_{x\in \X}\|x\|_\infty\bigr]\sqrt{e\log(d+1)\over m}.$$
 \item For any $0<\delta<1,$ with confidence at least $1-\delta,$ there holds
 \begeqn\label{eq:exm-1-2}\begin{array}{ll} \E_\bz(A_\bz)-\E(A_\bz) & \le {12\bigl[\sup_{x\in \X}\|x\|_F\bigr]\bigl[\sup_{x\in \X}\|x\|_\infty\bigr]\over r\gl}{\sqrt{e\log(d+1)\over m}} \\ & ~~~~+ \frac{2\bigl[\sup_{x\in \X}\|x\|_F\bigr]\bigl[\sup_{x\in \X}\|x\|_\infty\bigr]}{r\lambda}\sqrt{{2\ln \bigl({1\over \gd}\bigr)\over  m}}.\end{array}\endeqn

\item For any $0<\delta<1,$ with probability at least $1-\delta$ there holds
$$\begin{array}{ll} \mathscr{E}(f_\bz) & \leq  \E_{\bf z}(A_{\bf z})+
\frac{4\bigl[\sup_{x\in \X}\|x\|_F\bigr]\bigl[\sup_{x\in \X}\|x\|_\infty\bigr]}{\lambda r}{ \sqrt{2e\log(d+1)\over m}} \\ & ~~~~+\frac{2\bigl[\sup_{x\in \X}\|x\|_F\bigr]\bigl[\sup_{x\in \X}\|x\|_\infty\bigr]}{\lambda r} \sqrt{\frac{2\log\frac{1}{\delta}}{m}}.\end{array}$$
\end{enumerate}
\endexm
\begin{proof}The dual norm of the $(2,1)$-norm is the $(2,\infty)$-norm, which implies that $X_{*}=\dsup_{x,x'\in\X}\|x'x^T\|_{(2,\infty)}=\dsup_{x\in\X}\|x\|_F\dsup_{{x'\in\X}}\|x'\|_\infty$ and
$$\begin{array}{ll}&\EX_{{\bf z},\sigma}\dsup_{x\in\X} \Big\|\frac{1}{m}\dsum_{j\in\N_m}\sigma_j y_j x_j x^T\Big\|_{*}\leq \dsup_{{x\in\X}} \|x\|_F \EX_{{\bf z},\sigma} \dmax_{\ell\in\N_d}\Big|\frac{1}{m}\dsum_{j\in\N_m}\sigma_j y_j x_j^\ell\Big|\\
&\leq 2\dsup_{x\in\X}\|x\|_F\dsup_x \|x\|_\infty{\sqrt{e\log(d+1)\over m}},\end{array}$$
where the last inequality follows from estimation (\ref{eq:Rn-L1}).
We complete the proof by combining the above estimation with Theorems \ref{thm:generalresult} and \ref{thm:mainresult}.
\end{proof}

We briefly discuss the case of the trace norm regularization, i.e., $\|A\|=\|A\|_{\tr}.$  In this case, the dual norm of trace norm is the spectral norm defined, for any $B\in \mbS^{d\times d}$, by $\|B\|_\ast = \max_{\ell\in \N_d} \gs_\ell(B)$ where $\{\gs_\ell:\ell\in \N_d\}$ are the singular values of matrix $B.$ Observe, for any $u,v\in \R^d$, that  $\|uv^T\|_{\ast}=\|u\|_F\|v\|_F.$ Hence, the constant $X_{\ast}=\dsup_{x,x'\in\X}\|x'x^T\|_{\ast}=\dsup_{x\in\X}\|x\|_F^2.$ In addition,
\begeqn\label{eq:Rn-trace}\begin{array}{ll}\cR_m & = \EX_{{\bf z},\sigma}\dsup_{x\in\X} \big\|\frac{1}{m}\dsum_{j\in\N_m}\sigma_j y_j x_j x^T\big\|_{\ast}
\\  & = \EX_{{\bf z},\sigma}\dsup_{x\in\X} \|x\|_F \Big\|\frac{1}{m}\dsum_{j\in\N_m}\sigma_j y_j x_j \big\|_F    = \dsup_{x\in\X} \|x\|_F  \EX_{{\bf z},\sigma} \big\|\frac{1}{m}\dsum_{j\in\N_m}\sigma_j y_j x_j \big\|_F \\
 &  \leq \dsup_{x\in\X} \|x\|_F  \EX_{\bf z}\bigl(\EX_{\sigma} \big\|\frac{1}{m}\dsum_{j\in\N_m}\sigma_j y_j x_j \big\|^2_F\bigr)^{1\over 2}\\ &
= \dsup_{x\in\X} \|x\|_F \EX\bigl(\sum_{j\in\N_m} \|x_j\|_F^2\bigr)^{1\over 2 }\big / m.\end{array}\endeqn
Indeed, the above estimation for $\cR_m $ is optimal. To see this, we observe from \cite[Theorem 1.3.2]{pena} that $$ \bigl(\EX_{\sigma} \big\|\frac{1}{m}\dsum_{j\in\N_m}\sigma_j y_j x_j \big\|^2_F\bigr)^{1\over 2} \le \sqrt{2} \EX_{\sigma} \big\|\frac{1}{m}\dsum_{j\in\N_m}\sigma_j y_j x_j \big\|_F.$$
Combining the above fact with (\ref{eq:Rn-trace}), we can obtain
$$\beg{array}{ll} \cR_m &  = \dsup_{x\in\X} \|x\|_F  \EX_{{\bf z},\sigma} \Big\|\frac{1}{m}\dsum_{j\in\N_m}\sigma_j y_j x_j \Big\|_F \ge {1\over \sqrt{2}} \dsup_{x\in\X} \|x\|_F  \EX_{{\bf z}} (\EX_{\sigma} \big\|\frac{1}{m}\dsum_{j\in\N_m}\sigma_j y_j x_j \big\|^2_F)^{1\over 2} \\ &
=  {1\over  m \sqrt{2}} \dsup_{x\in\X} \|x\|_F \EX \bigl( \sum_{j\in\N_m}\|x_j\|^2\bigr)^{1\over 2}. \end{array} $$
Hence, the estimation (\ref{eq:Rn-trace}) for $\cR_m$  is optimal up to the constant ${1\over \sqrt{2}}.$ Furthermore, ignoring further estimation ${1\over  m }\EX \bigl( \sum_{j\in\N_m}\|x_j\|^2\bigr)^{1\over 2} \le {1\over  \sqrt{m} }\dsup_{x\in\X} \|x\|_F  $, the above estimations mean that the estimation for   $\cR_m$ in the case of trace-norm regularization are the same as the estimation (\ref{eq:Rn-Fro}) for the Frobenius norm regularization. Consequently, the generalization bounds for similarity learning and the relationship between similarity learning and the linear SVM are the same as those stated in Example \ref{exm:fro}. It is a bit disappointing that there is no improvement when using the trace norm. The possible reason is that the spectral norm of $B$ and the Frobenius norm of $B$ are the same when $B$ takes the form $B=xy^T$ for any $x,y\in\R^d.$

We end this section with a comment on an alternate way to estimate the Rademacher average $\cR_m$.   Kakade \cite{Kakade-a,Kakade-b} developed elegant techniques for estimating Rademacher averages for linear predictors. In particular, the following theorem was established:

\begth\label{thm:kakade} \emph{(\cite{Kakade-a,Kakade-b})} Let $\mathcal{W}$ be a closed convex set and let $f: \mathcal{W} \to \R$ be a $\gb$-strongly convex with respect to $\|\cdot\|$ and assume that $f^\ast(0) =0$.   Assume $\mathcal{W} \subseteq \{w: f(w)\le f_{\max}\}.$ Furthermore, let $\mathcal{X} = \{x: \|x\|_\ast \le X\}$ and $\mathcal{F} = \{w \to \langle w, x\rangle: w \in \mathcal{W}, x\in \mathcal{X}\}.$ Then, we have
$$R_n(\mathcal{F}) \le X \sqrt{ 2 f_{\max}\over \gb n }.$$
\endth
To apply Theorem \ref{thm:kakade},  we rewrite  the Rademacher average $\cR_m$ as
\begeqn\label{eq:Rn-form}\begin{array}{ll}\cR_m  &  = \EX_{\bz,\sigma}\Big[\dsup_{\tilde{x}\in \X}\Big\|\frac{1}{m}\dsum_{i\in\N_m}\sigma_i y_i x_i \tilde{x}^T \Big\|_{\ast}\Big] \\ &
= \EX_{\bz,\sigma} \Big[\dsup_{\tilde{x}\in \X}\sup_{\|A\|\le 1, A \in \mbS^{d}} \langle \frac{1}{m}\dsum_{i\in\N_m}\sigma_i y_i x_i \tilde{x}^T, A\rangle\Big]\\ & = \EX_{\bz,\sigma} \Big[\dsup_{\tilde{x}\in \X}\sup_{\|A\|\le 1, A \in \mbS^{d}} \langle \frac{1}{m}\dsum_{i\in\N_m}\sigma_i y_i x_i,  A \tilde{x}\rangle\Big].
 \end{array} \endeqn
Now let $\mathcal{W}:  = \{w \to \langle w, x \rangle, w= A \tilde{x}, \|A\|\le 1, A \in \mbS^{d\times d}\}.$  Let us consider  the sparse $L^1$-norm defined, for any $A\in \mbS^{d\times d}$, by $\|A\|_1 = \sum_{k,\ell\in\N_d} |A_{k\ell}|.$ In this case, we observe that $\|w\|_1 \le  \|A\|_1 \|\tilde{x}\|_\infty\le   \|A\|_1 \sup_{x\in \X}\|x\|_\infty\le \sup_{x\in \X}\|x\|_\infty.$  Let $f(w) = \|w\|^2_q$ with $q = {\log d \over \log d -1}$ which is ($1\over \log d$)-strongly convex with respect to the norm $\|\cdot\|_1.$ Then, for any $w\in \mathcal{W}$, we have that $\|w\|^2_q \le \|w\|_1^2 \le\sup_{x\in \X}\|x\|_\infty^2.$ Combining these observations with (\ref{eq:Rn-form}) allows us to obtain the estimation
$\cR_m  \le \sup_{x\in \X}\|x\|_\infty^2 \sqrt{2 \log d \over m}.$  Similarly, for the $(2,1)$-mixed norm $\|A\|_{(2,1)} = \sum_{k\in\N_d}(\sum_{\ell\in\N_d} |A_{k\ell}|^2)^{1/2}$, observe that $\|w\|_1 \le  \|A\|_{(2,1)} \|\tilde{x}\|_F\le   \sup_{x\in \X}\|x\|_F.$ Applying Theorem \ref{thm:kakade} with $f(w) = \|w\|^2_q$ ($q = {\log d \over \log d -1}$) again, we will have the estimation $ \cR_m  \le \sup_{x\in \X}\|x\|_F\sup_{x\in \X}\|x\|_\infty  \sqrt{2 \log d \over m}.$  Hence, the estimations for the above two cases are similar to our estimations in the above examples. Our estimation is more straightforward by directly using the Khinchin-type inequality in contrast to the advanced convex-analysis techniques used in Kakade et al. \cite{Kakade-a,Kakade-b}.

However, for the case of trace-norm regularization (i.e., $\|A\|=\|A\|_{\tr}.$), one would expect, using the techniques in \cite{Kakade-a,Kakade-b}, that the estimation for $\cR_m$ is the same as that in the case for the sparse $L^1$-norm. The main hurdle for such result is the estimation of $\|w\|_1 = \|A\tilde{x}\|_1$ by the trace-norm of $A.$  Indeed, by the discussion following our estimation (\ref{eq:Rn-trace}) directly using Khinchin-type inequality, we know that our estimation (\ref{eq:Rn-trace}) is optimal.  Hence, one can not expect the estimation for $\cR_m$ for the case for trace-norm regularization is the same as that in the case for sparse $L^1$-norm regularization in our particular case of similarity learning formulation (\ref{algorithm}).

\section{Conclusion}\label{sec:conclusion}

In this paper, we considered a  regularized similarity learning formulation (\ref{algorithm}). Its generalization bounds were established for various matrix-norm  regularization terms such as the Frobenius norm, sparse $L^1$-norm, and mixed $(2,1)$-norm. We proved the generalization error of the linear separator based on the learnt similarity function can be bounded by the derived generalization bound of similarity learning. This guarantees the goodness of the generalization of similarity learning (\ref{algorithm}) with general matrix-norm regularization and thus the classification generalization of the resulting linear classifier.  Our techniques using the Rademacher complexity \cite{BM} and the important Khinchin-type inequality for the Rademacher variables allow us to obtain  new bounds for similarity learning with general matrix-norm regularization terms.

There are several possible directions for future work. Firstly, we may consider similarity algorithms with general loss functions. It is expected that under some convexity conditions on the loss functions, better results could be obtained. Secondly, we usually focus on the excess misclassification error when considering classification problems. Hence, in the future, we would like to consider the theoretical link between the generalization bounds of the similarity learning and the excess misclassification error of the classifier built from the learnt similarity function.

\section*{Acknowledgement:}{We are grateful to the referees for their invaluable comments and suggestions on this paper. This work was supported by the EPSRC under grant EP/J001384/1. The corresponding author is Yiming Ying.}

\section*{Appendix}

In this appendix, the following facts are used for establishing
generalization bounds in section \ref{sec:generalizationbound} and section \ref{sec:classification}.

\begin{definition} We say the function $f:\displaystyle\dprod_{k=1}^m \Omega_k \rightarrow
\mathbb{R}$ with bounded differences $\{ c_k \}_{k=1}^m $ if, for
all $1\le k \le m $, $$\begin{array}{ll} \dmax_{z_1,\cdots
,z_k,z^\prime_{k}\cdots ,z_m} & |  f(z_1, \cdots
,z_{k-1},z_k,z_{k+1}, \cdots , z_m)\\ & -f(z_1, \cdots
,z_{k-1},z^\prime_k,z_{k+1}, \cdots , z_m) |\le c_k\end{array}
$$
\end{definition}

\begin{lemma}\label{lem:McD}  (McDiarmid's inequality \cite{McD}) Suppose $f:\displaystyle\dprod_{k=1}^m \Omega_k \rightarrow
\mathbb{R}$ with bounded differences $ \{ c_k \}_{k=1}^m $ then ,
for all $\epsilon>0 $, there holds
$$ {\bf Pr}_{\bf z} \biggl\{ f(\bz) -\mathbb{E}_{\bf z}f({\bf z})\ge \epsilon\biggr\}\le
 e^{- \frac{2\epsilon^2}{\sum_{k=1}^m c_k^2}}.$$
\end{lemma}
We need the following contraction  property of the Rademacher
averages which  is essentially implied by Theorem 4.12 in Ledoux and
Talagrand \cite{LT},  see also \cite{BM,KPan}.

\begin{lemma}\label{lem:contr-prop}Let $F$ be a class of uniformly bounded real-valued
functions on $(\gO,\mu)$ and $m\in\N$. If for each $i\in \{1,
\ldots, m\}$, $\phi_i: \R \to \R$ is a function
having a Lipschitz constant $c_i$, then for any $\{x_i\}_{i\in\N_m}$,
\begeqn \EX_\epsilon\Big( \dsup_{f\in F} \dsum_{i\in\N_m}
\epsilon_i \phi_i(f(x_i)) \Big)
 \le 2   \EX_\epsilon\Big(  \dsup_{f\in F}
 \dsum_{i\in\N_m} c_i\epsilon_i
f(x_i)  \Big).\label{contr}\endeqn

\end{lemma}
Another important property of the Rademacher average which is used in the proof of the generalization bounds of the similarity learning is the following Khinchin-type inequality, see e.g. \cite[Theorem 3.2.2]{pena}.
\begin{lemma}\label{Khinchin}
For $n\in \N,$ let $\{f_i\in \R: i\in\N_n\},$ and $\{\sigma_i: i\in\N_n\}$ be a family of i.i.d. Rademacher random variables. Then, for any $1<p<q<\infty$ we have
$$\Big(\EX_\sigma\Big|\dsum_{i\in\N_n}\sigma_if_i\Big|^q\Big)^{1\over q}\leq\Big(\frac{q-1}{p-1}\Big)^{1\over2}\Big(\EX_\sigma\Big|\dsum_{i\in\N_n}\sigma_if_i\Big|^p\Big)^{1\over p}.$$
\end{lemma}

\end{document}